\documentclass[a4paper]{llncs}

\usepackage{makeidx}  
\usepackage{listings}
\usepackage[pdftex]{graphicx}
\usepackage{array}
\usepackage{url}
\newcommand{\keywords}[1]{\par\addvspace\baselineskip
\noindent\keywordname\enspace\ignorespaces#1}
\usepackage{subfig}
\begin{document}
\mainmatter              
\title{Optimal Approach for Image Recognition using Deep Convolutional Architecture}
\titlerunning{\ }  
%
\author{Parth Shah\inst{1} \and Vishvajit Bakrola\inst{2} \and Supriya Pati\inst{3}
}
\authorrunning{\ }   
%
\tocauthor{Parth Shah, Vishvajit Bakarola, Supriya Pati}
\institute{C.G Patel Institute of Technology, Uka Tarsadia University, Bardoli, India\\\inst{1} \email{parthpunita@yahoo.in},  \inst{2} \email{vishvajit.bakrola@utu.ac.in},\\ \inst{3} \email{supriya.pati@utu.ac.in}\\
}
\maketitle              

\begin{abstract}        
In the recent time deep learning has achieved huge popularity due to its performance in various machine learning algorithms. Deep learning as hierarchical or structured learning attempts to model high level abstractions in data by using a group of processing layers. The foundation of deep learning architectures is inspired by the understanding of information processing and neural responses in human brain. The architectures are created by stacking multiple linear or non-linear operations. The article mainly focuses on the state-of-art deep learning models and various real world applications specific training methods. Selecting optimal architecture for specific problem is a challenging task, at a closing stage of the article we proposed optimal approach to deep convolutional architecture for the application of image recognition.

\end{abstract}
\section{Introduction}
In any artificial intelligence problem we require two main things. First, we need to identify and extract right set of features that represent the problem. Second, we need to have an algorithm that takes these extracted features and provides predicted outputs. Identifying right set of features itself is a challenging task especially when we are dealing with the images. The solution of this problem is to allow machines to learn from their own experience instead of fixed rules and to understand concepts in terms of hierarchy of simpler concepts. If we create a graph that shows how these concepts are stacked on each other then that resulting graph becomes deep with high number of layers. This is why we call this approach deep learning. In deep learning, we normally use deep neural networks where each neuron in same layer maps different features and each layer will combine features of previous layer to learn new shapes.
But this brings new challenge of choosing the perfect strategy for implementing deep learning architecture as accuracy and time required for training depends on it.

 Recently, very deep convolutional neural networks are in main focus for image or object recognition. For the task of image recognition several different models like LeNet, AlexNet, GoogLeNet, ResNet and Inception-ResNet etc. are available. Most of these models are result of ImageNet Large Scale Visual Recognition Challenge (LSVRC) and MSCOCO Competition, which is a yearly competition where teams from around the globe compete for achieving best accuracy in image recognition task. In ImageNet LSVRC, evaluation criteria are top - 1 and top - 5 error rate, where the top - N error rate is the fraction of test images for which the correct label is not among the N labels considered most probable by the model. In addition to that we can judge any deep neural network based on computation cost, memory it requires to execute, etc.
  Selecting best appropriate model from that is tricky task. It depends on size of input, type of input as well available resources.  In this paper, we have analyzed effect of different training methods on these models and evaluated performance on different size of dataset. We have also discussed the benefits and trade offs of increasing number of layers.


In Section 2 of this paper we have presented literature review of different deep learning models for image recognition in detail. In Section 3, we have described different training methodologies for deep neural network architectures. Section 4 describes about implementation environment used for implementing various model described in literature review. Comparison of these model under various scenario are presented in Section 5. In Section 6, we conclude the finding of this paper.

\section{Literature Review}
First concept of deep learning was introduced way back in early 80's when computers were not in even day to day usage. In 1989, Yann LeCun successfully demonstrated deep convolutional neural network called LeNet for task of hand written character recognition. But it was not further developed because there was not enough data and high computation power available at that time which was required by deep neural netwoks. This slowed down the research in area of deep learning. The new wave of research started only after Alex Krizhevsky successfully demonstrated use of deep convolutional neural networks by beating traditional object recognition methods in ImageNet LSVRC-2012 by large margin in 2012. After that year on year new deep neural networks were introduced like GoogLeNet, ResNet, Inception-ResNet etc. Each architecture was designed to have more accuracy than its precedent. We have covered all these models in brief in this section.

\subsection{LeNet}

LeNet consist of 5 convolution layers for feature extraction and object detection. Before LeNet, people used methods that requires feature vector to be provided to algorithm. These feature vector needed to be handcrafted from knowledge about the task to be solved. This problem is solved in LeNet by using convolutional layers as a feature extractor \cite{726791}. These convolutional layer's weight are learnable parameters which we can use during training process. Due to usage of convolution layers in LeNet it requires high computation power. In order to tune weight of these convolutions higher number of training dataset is required. These two were the main limitations of LeNet when it was introduced. This prevented further development in deep learning in early 80's.

\subsection{AlexNet}
One of the major breakthrough in the area of deep learning was achieved when Alex Krizhevsky successfully demonstrated use of convolutional neural networks by beating traditional object recognition methods in ImageNet LSVRC by large margin in 2012. AlexNet's architecture was based on concept established by LeNet. AlexNet uses total 8 different hidden layers. From this eight layers, first five layers are convolution layers and other three layers are fully connected layers. Here these convolution layers are used for feature extraction task. Lower layer will extract basic features like edges. As we go to higher level it combines shapes from lower layer and identify shapes. Output of last fully connected layer is fed as input to 1000 way softmax layer which act as output layer which represents 1000 class of ImageNet dataset \cite{deng2009imagenet}. Softmax layer will output probability of each output class between 0 to 1, where 0 means object is not present in an image while 1 means object is present in an image. Demonstrated AlexNet model at ILSVRC-2012 used GPU for meeting computation needs of these convolution layers. Alexnet had directly reduced top-5 error rate of 26\% of 2011 ImageNet winner to 16\% which was more than 10\% improvement in single year. One of the main reason of this huge performance improvement was instead of hard coding of features, it had extracted featured automatically using deep convolutional layers like LeNet \cite{krizhevsky2012imagenet}.

\subsection{GoogLeNet}
Although AlexNet improves accuracy greatly compared to traditional architectures, it was not able to provide human like accuracy because only 8 layers were not able to extract all features needed for identifying all 1000 classes in ImageNet dataset. Based on architecture of AlexNet, Szegedy et al. developed new deep convolutional neural network architecture called GoogLeNet in 2014 \cite{szegedy2015going}. GoogLeNet took concept of building hierarchy of feature identifier from AlexNet and stacked layers in form of inception modules. It uses the concept of network in network strategy \cite{DBLP:journals/corr/LinCY13} where the whole network is composed of multiple local network called `inception' module. These inception modules consists of 1x1, 3x3 and 5x5 convolutions.  All convolutional layers in GoogLeNet are activated by use of rectified linear unit. GoogLeNet has 3 times more layers compared to AlexNet. Architecture of GoogLeNet is 27 layers deep (more than 100 layers if we count layers in inception module separately). Inception module is designed such that it provides better result as compared to directly stacking layer on one another like in AlexNet. The network was designed with computational efficiency and practicality in mind, so that inference can be run on any devices including those with limited computational resources and low-memory footprint. GoogLeNet has achieved around 6.67\% top-5 error rate and won the ImageNet LSVRC-2014 competition \cite{szegedy2015rethinking}.

\subsection{ResNet}
Even with the optimized architecture of GoogLeNet, deep neural network remained difficult to train. In order to increase the ease of training and accuracy the concept of residual connection was added to deep neural network. In 2015, ResNet architecture proposed by He et al. achieved super human accuracy of just 3.57\% error rate using residual connections in ImageNet LSVRC-2015 competition and MSCOCO competition 2015 \cite{DBLP:journals/corr/HeZR015}. The reason for adding residual connection was that when deeper network starts converging, problem of degradation occurs where accuracy degrades rapidly after some point in training. In order to solve this problem, instead of hoping that each stacked layers directly fit a desired underlying mapping, it explicitly lets these layers fit to a residual mapping using newly added residual connections. Architecture of ResNet was based on VGGNet \cite{DBLP:journals/corr/SimonyanZ14a} which was the runners up of ImageNet LSVRC-2013 competition. VGGNet was composed of fixed size convolution layer with varying number of deepness of architecture. In ResNet, between each pair of 3x3 convolutional layer a shortcut residual connection was added. Authors have proposed different architecture of ResNet having 20, 32, 44, 56, 110, 152 and 1202 layers. But as we increase number of layers computation complexity increases due to higher number of convolution operations.

\subsection{Inception-ResNet}
In order to optimize architecture of deep neural networks, Szegedy at el. has introduced residual connection's concept from ResNet into Inception module in their GoogLeNet architecture and proposed Inception-ResNet architecture in 2016. It helps to keep performance of network while accelerating training of network using residual connection. Inception-ResNet was able to achieve error rate of just 3.08\% over ILSVRC dataset \cite{szegedy2016inception}. In Inception-ResNet, shortcut connection was added between each inception module. Inception-Resnet used two simple convolution model of ResNet with single Inception module. Authors have proved that in addition to increasing model size, residual connection also increase training.

\section{Training Methodologies of Deep Neural Networks}
Once architecture of neural network is defined we need to train it so that it can learn the given problem. Training can be done in various way but its main goal is to map given input data to its appropriate given output value. Once training is done we can save final updated mapping and use it while performing inference on test data. Two of the approaches that are highly used in training for deep neural networks are as follows:

\subsection{Training from Scratch}

This is the most common method for training neural networks. In this method at the start of training, correct class for each test data is known. Then we will initialize weights of all layers randomly. After initialization we will try to map these class label with actual input. We will adjust weight of each neuron such that model will learn to predict actual class label. Using this approach we will train all the available neurons in networks such that model will learn to output correct label. As this approach requires to update each neuron's weight, it requires large dataset for providing higher accuracy.

\subsection{Training using Transfer Learning}

Generally traditional algorithms were developed to train on specific task as it required to extract features manually, but with the introduction of deep learning, process of feature extraction become part of neural network itself. This introduces the new window of opportunity of generalized architecture that can deal with more than one type of problem. That's where the concept of transfer learning is introduced. Transfer learning provides the advance way of learning in machine learning algorithms. Transfer learning is the method of improvement of knowledge in a new task using knowledge transfer from previously learned similar task \cite{transfer_learning}.
 The most common method for applying transfer learning is to only train neuron of final layer keeping neurons of all other layers fixed. This approach will greatly speedup process of learning.
Generally in deep learning, due to higher number of layers there are larger number of trainable weights are there. But in most of the cases only weight in last layers are deciding factor for output generation. This makes transfer learning best suitable training method when we want to train model with higher number of layers but we get only small dataset due to some limitations.

\section{Implementation}
For evaluating performance of different models of deep neural network architecture we have implemented some of the best models based on accuracy for object recognition and compared its performance based on different dataset size, different number of labels and different training methods.

Dataset and implementation environment we used for implementing these image recognition models are discussed in details in this section.
\subsection{Datasets}
Different dataset we used for testing are as follows:

\noindent\textbf{MNIST Handwritten Digit Dataset:}
The MNIST database is a collection of handwritten digits that is created by National Institute of Standard and Technology. It contains 28x28 pixel grayscale image for english numerical digits \cite{726791}. The standard MNIST
database contains 60,000 training images and 10,000 testing images. We referred it as MNIST Full dataset throughout this paper. In addition to this we have created two another dataset. From which one dataset contain subset of around 12000 image called MNIST Small and another contains total 140000 images called MNIST Inverted dataset.

\noindent\textbf{Flowers Dataset:}
Flower dataset consists of different flower images that is collected from the internet having creative commons license. It has only 5 different labels and each label contains around 650 images.

\noindent\textbf{Yale Face Dataset:}
Yale Face dataset is collection of face images captured under different lighting conditions and different angels for multiple persons. It has around 40 different faces where each face has around 60 different images \cite{Georghiades:2001:FMI:378040.378083}.

\subsection{Implementation Environment}
Implementation and training of these models are done using tensorflow deep learning framework using two different machine setups. They are as follows:
\noindent\textbf{Setup 1:}
First machine is an AWS GPU instance with Xeon E5 processor with 8 core and NVIDIA K520 GPU having 3072 CUDA cores and 8GB GPU memory and 15GB RAM.

\noindent\textbf{Setup 2:}
Second machine is dedicated hardware with Intel Xeon E3 processor with 12 cores and 32 GB RAM.
\section{Comparison}

\subsection{Training with MNIST dataset}
In first experiment we have trained LeNet, AlexNet and GoogLeNet using MNIST dataset over hardware setup 1.
Before performing actual training of the network we need to preprocess them. In preprocessing we have converted all image files to single file such that it can be used for batch processing.

\begin{figure}[h]
\centering{}
\vspace{-15px}
\includegraphics[width=12cm,height=3cm]{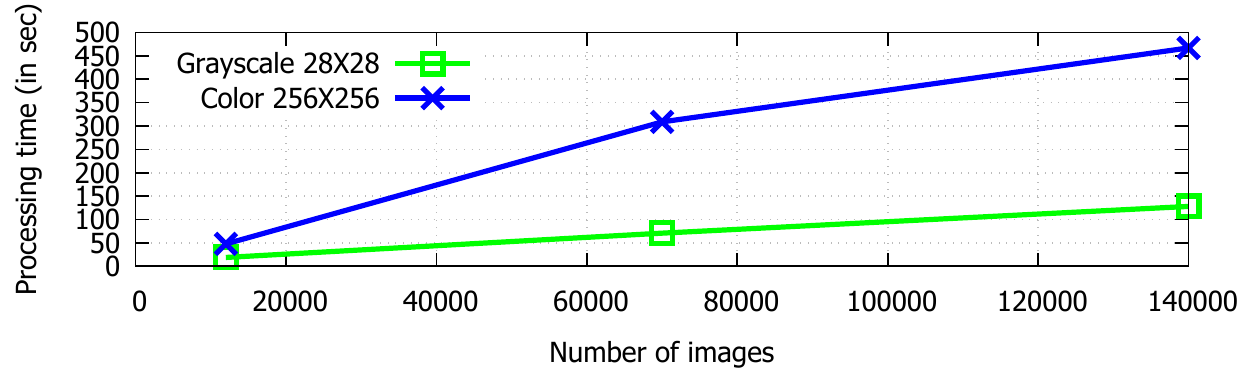}\caption{Preprocessing cost for different size of dataset}\label{mnist_preprosess}
\end{figure}

As we can see from Fig. \ref{mnist_preprosess}, time require for preprocessing is dependent on number of images in dataset as well as number of channels in image.

\begin{figure}[h]\centering
\vspace{-15px}
\subfloat[MNIST sample digits\cite{726791}]{
\includegraphics[width=4.5cm,height=2.5cm]{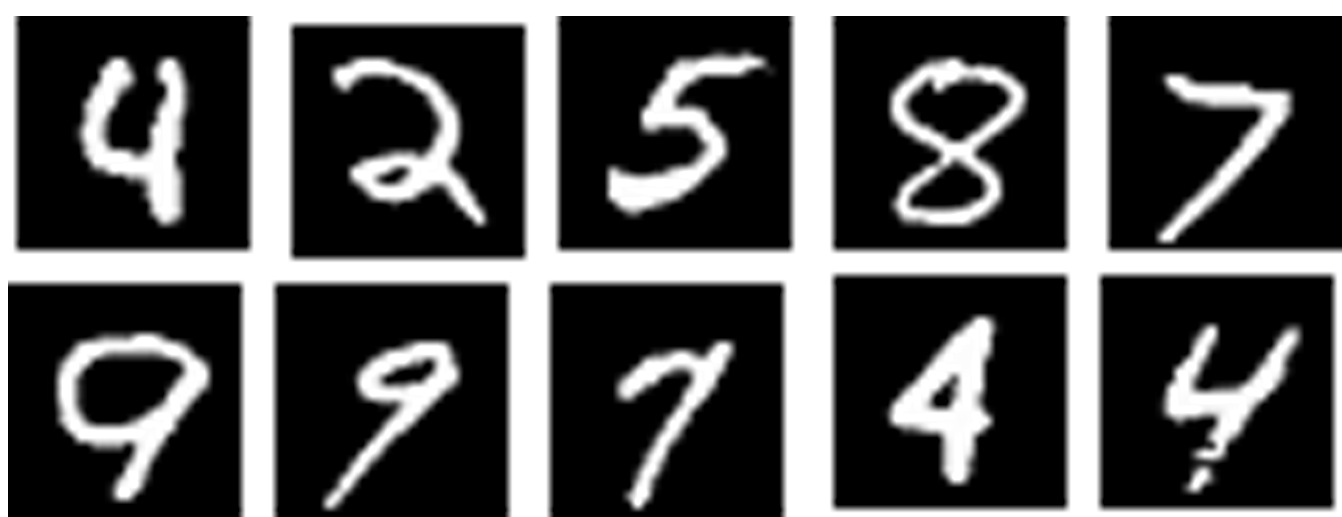}}
\subfloat[Results]{
  \includegraphics[width=7cm,height=2.5cm]{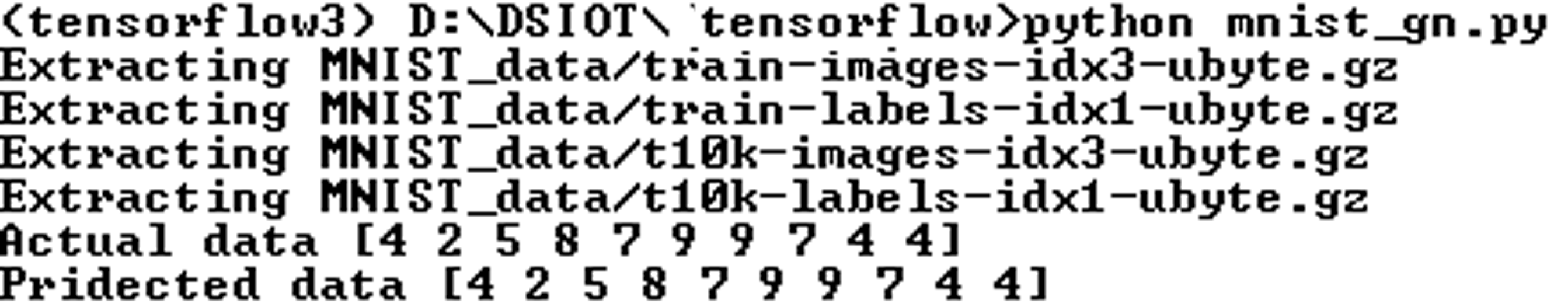}}
\caption{Handwritten digit recognition example}\label{digitidentification}
\end{figure}

After data is preprocessed we apply it to train our models. Each model is trained for 10 epoch with base learning rate of 0.01 and learning rate will decrease with every 3 steps in factor of 0.01 to increase the efficiency of training. We have measured Top-1 accuracy, Top-5 accuracy and training loss for our training in Table \ref{mnist_perf}. As Top-5 accuracy was introduced as a criteria for ImageNet LSVRC competition in 2013, it is only available in models that are developed after that. So for other network we have used NA in table.
\begin{table}[h]
\centering{}
\vspace{-15px}
\caption{Performance of different models on MNIST}
\label{mnist_perf}
\begin{tabular}{>{\centering}m{3cm}>{\centering}m{1.2cm}>{\centering}m{1.2cm}>{\centering}m{1.5cm}>{\centering}m{1.5cm}>{\centering}m{1.8cm}>{\centering}m{1.8cm}}
\hline
Dataset & MNIST SMALL & MNIST FULL  & MNIST SMALL & MNIST FULL & MNIST SMALL & MNIST FULL\tabularnewline
\hline
Model & LeNet & LeNet & AlexNet & AlexNet & GoogLeNet & GoogLeNet\tabularnewline
Layers & 5 & 5  & 8  & 8 & 22 & 22\tabularnewline
Accuracy in Top-1 & 96.80\% & 98.86\% & 92.22\% & 99.24\% & 85.50\% & 98.26\%\tabularnewline
Accuracy in Top-5 & NA & NA & NA  & NA & 98.5372\% & 99.96\%\tabularnewline
Loss & 0.105 & 0.040 & 0.256 & 0.024 & 0.438 & 0.061 \tabularnewline
Time & 50s & 45s &  454s & 2010s & 1167s & 9300s\tabularnewline
\hline
\end{tabular}
\end{table}
\begin{figure}[h]
  \centering
  \vspace{-15px}
  \includegraphics[width=13cm,height=3.5cm]{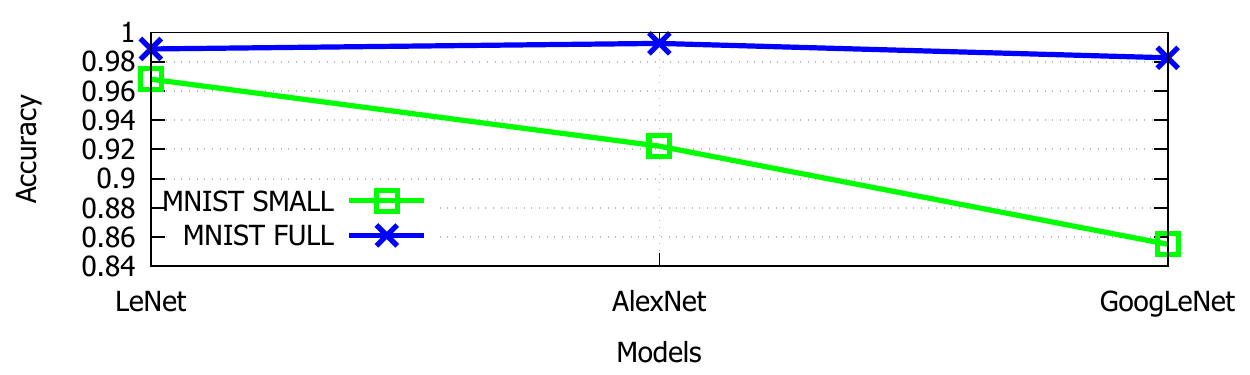}
  \caption{Accuracy for different models}\label{mnist_accuracy}
\end{figure}
\vspace{-15px}

Fig. \ref{digitidentification} shows how trained model can easily identify handwritten digits with high accuracy.
As we can see from Fig. \ref{mnist_accuracy}, when training set is large then performance improves. Similarly, increasing number of layers also increases accuracy with side effect of increasing training time. Other thing we can notice is that increasing number of layers on smaller dataset decreases performance as model also picks up unwanted noise present in data set as a feature. It is also known as overfitting problem.

\subsection{Training with Flowers dataset}

As we can see from previous experiment, for smaller dataset performance degrades compared to larger dataset. In order to solve this degradation problem, transfer learning method is used. In this experiment, we have trained GoogLeNet model from scratch and also trained GoogLeNet and Inception-ResNet using transfer learning on Flowers dataset. For training all three model we have used hardware setup 2 which is described in previous section.

\begin{figure}[h]\centering
\vspace{-15px}
\subfloat[Daisy\cite{deng2009imagenet}]{
\includegraphics[width=3.5cm,height=2.5cm]{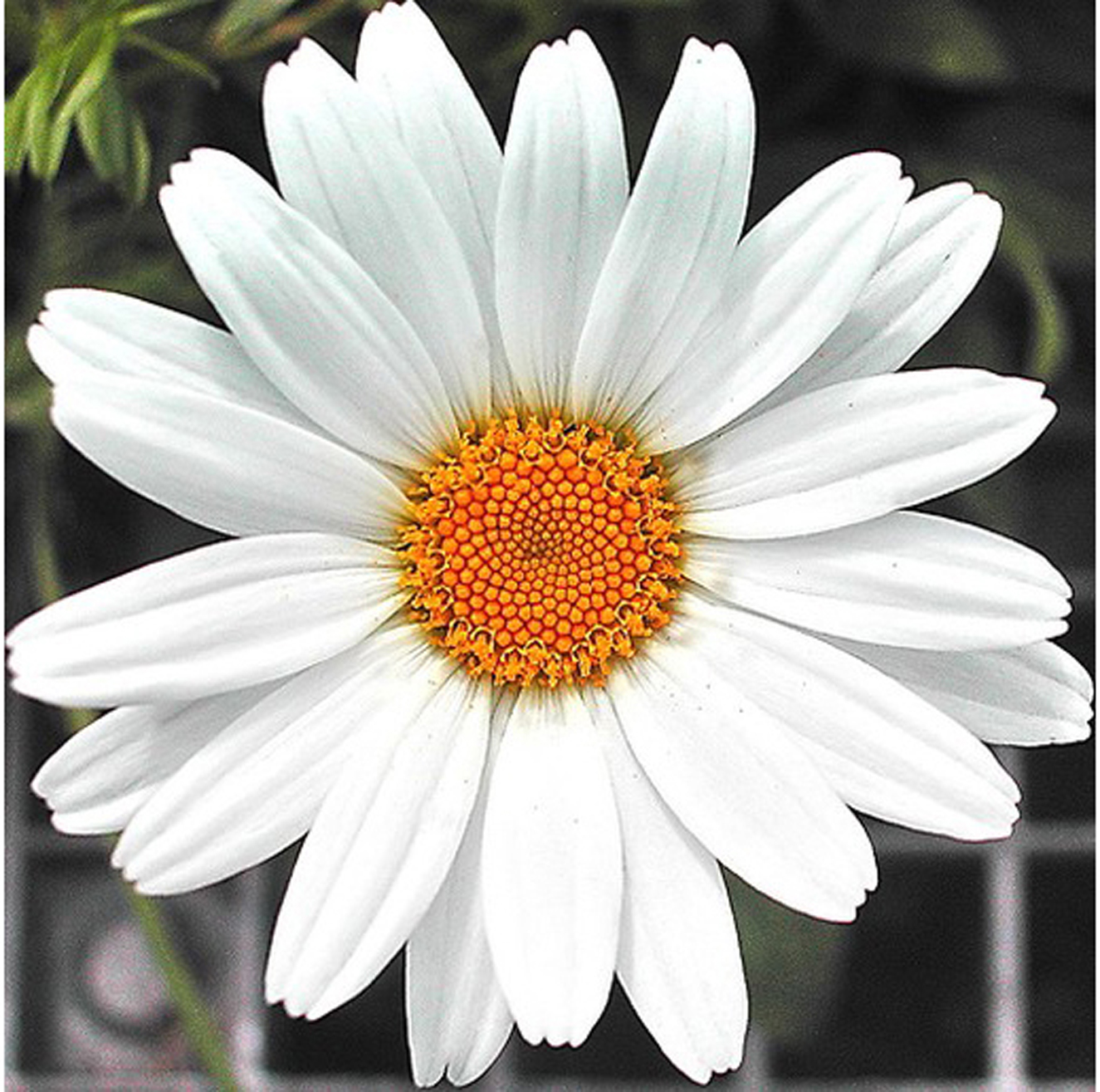}}
\subfloat[Results]{
  \includegraphics[width=6.5cm,height=2.5cm]{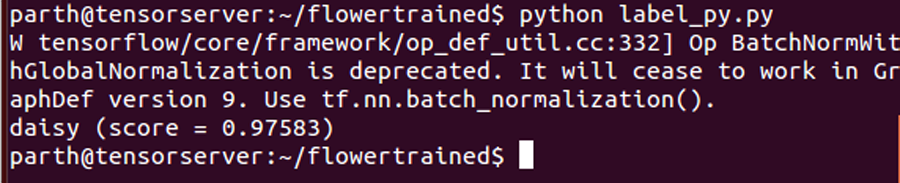}}
\caption{Daisy identification example}\label{daisyidentification}
\end{figure}

Simple training is done for GoogLeNet using 40000 epoch in case of learning from scratch, while for transferred leaning on GoogLeNet and Inception-ResNet, 10000 training epoch is used with base learning rate of 0.01 and learning rate will decrease every 30 step with factor of 0.16 for optimizing learning process. Simple example of trained model using transfer learning is shown in Fig. \ref{daisyidentification}.

\vspace{-15px}

As seen from results in Table \ref{scratch_vs_learning_table}, using transfer learning we can achieve better performance compare to training from scratch as in transfer learning it preserves the features it extracted when it was previously trained on larger dataset. This greatly decrease training time required for training maintaining accuracy.
\begin{table}[h]
\centering{}
\vspace{-15px}
\caption{Comparison of learning from scratch vs transfer learning}
\label{scratch_vs_learning_table}
\begin{tabular}{>{\centering}m{3cm}>{\centering}m{2cm}>{\centering}m{3cm}>{\centering}m{3cm}}
\hline
Training Method & From scratch & Transfer Learning & Transfer Learning \tabularnewline
\hline
Model & GoogLeNet & GoogLeNet & Inception-ResNet\tabularnewline
Accuracy (Top-1) & 86.7\% & 90.6\% & 93.7\% \tabularnewline
Number of epoch & 40000 & 10000 & 10000 \tabularnewline
Training Time & 245 hours & 12 hours & 14 hours\tabularnewline
\hline

\end{tabular}
\end{table}
\vspace{-15px}

\subsection{Training with Yale Face dataset}
Normally in real world usage scenario of any image recognition problem inputs are taken from different image capturing devices which are fixed at particular place like security cameras. It is not always possible to have full object captured by camera everytime. Our model should be capable of identifying an object even when only partial object is present in image.
In order to check how model performs in case of incomplete or partial images, we have first trained model with Yale Face dataset and then tested it with image containing partial objects. This experiment was performed on hardware setup 2 using transferred learning approach. Base learning rate of 0.01 is used for initialization of training. Learning rate will decrease every 30 step with factor of 0.16.

\begin{figure}[h]
  \centering
  \vspace{-15px}
  \includegraphics[width=10cm,height=3cm]{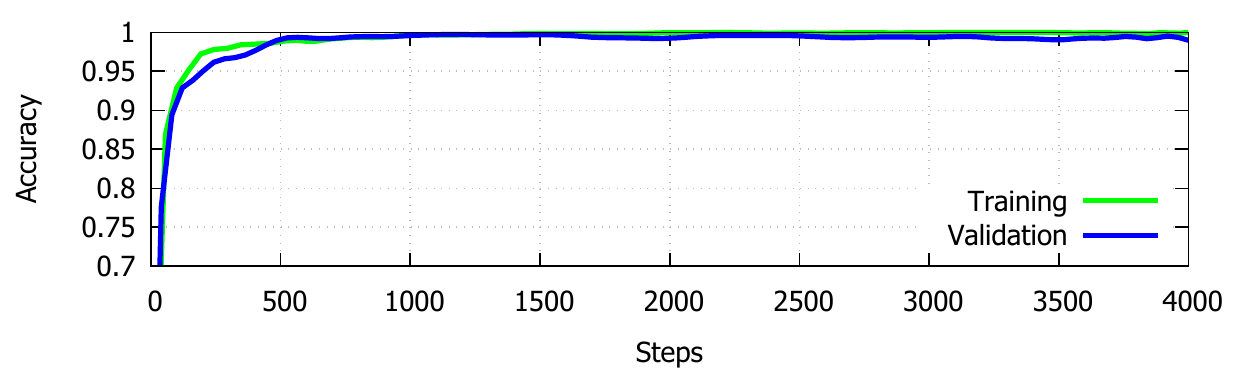}
  \caption{Accuracy of training in Yale Face dataset}\label{head_graph}
\end{figure}
\vspace{-15px}
Fig. \ref{head_graph} shows overall training accuracy and validation accuracy for Yale Face dataset. As you can see after around 1000 epoch, improvement of accuracy is minimal.

\begin{figure}[h]\centering
\vspace{-15px}
\subfloat[30\% Face \cite{Georghiades:2001:FMI:378040.378083}]{
\includegraphics[width=3.5cm,height=3cm]{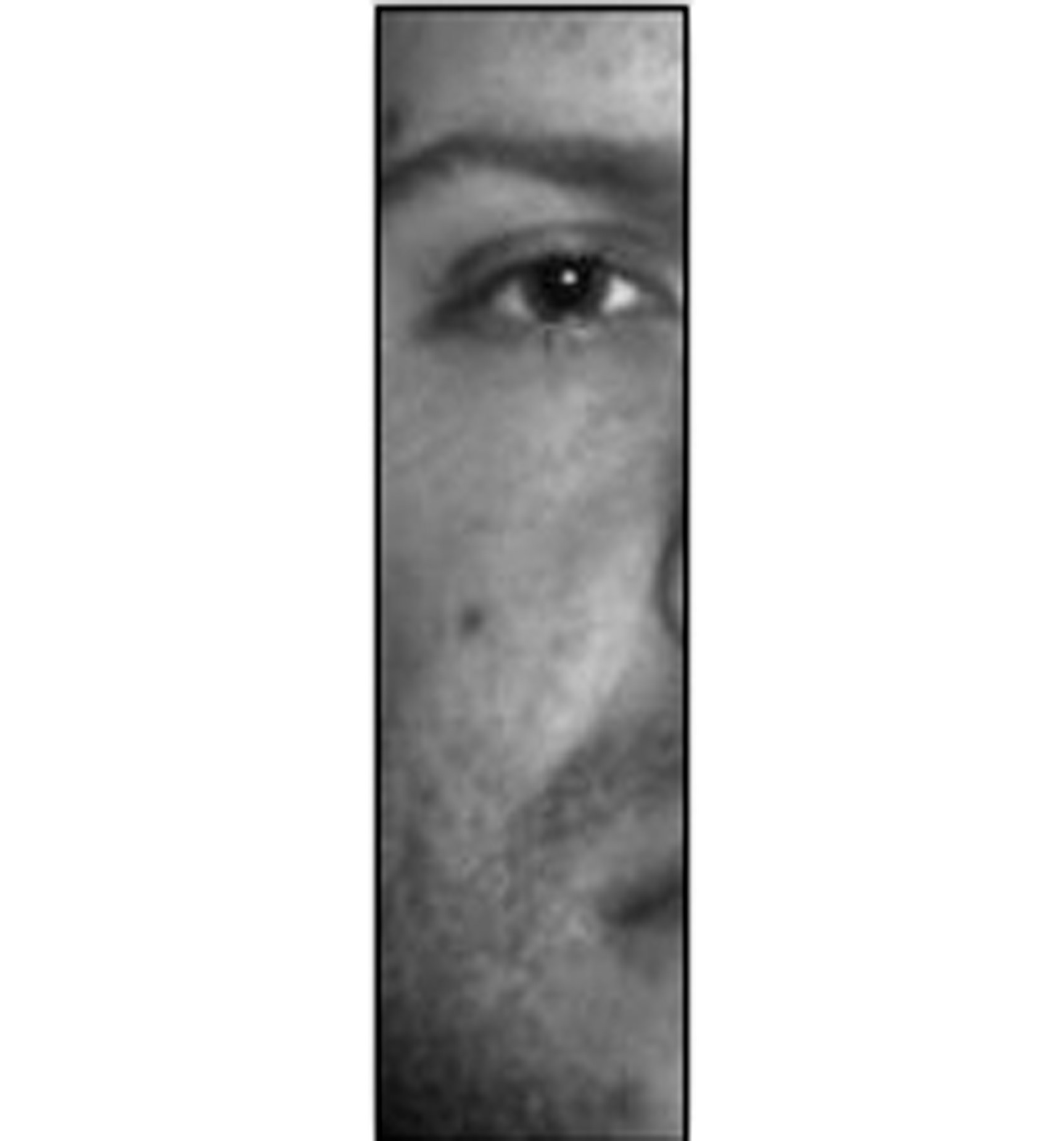}}
\subfloat[Successful Identification]{
  \includegraphics[width=7.5cm,height=3cm]{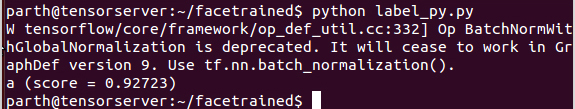}}
\caption{Face identification in case of partial image}\label{incomplete_img}
\end{figure}

%
\vspace{-15px}
 As seen from Fig. \ref{incomplete_img} that model provides correct results even when there is only 30\% of face present in image. This is due to nature of deep learning that extracted minute features that were previously not possible with hardcoded features of traditional approaches.
\begin{table}[h]
\centering{}
\vspace{-15px}
\caption{Results in case of partial object in image}
\label{yale_face_table}
\begin{tabular}{>{\centering}m{4cm}>{\centering}m{5cm}}
\hline
Amount of object in image & Average output of Softmax layer\tabularnewline
\hline
10\% & 0.92642 \tabularnewline
20\% & 0.84098 \tabularnewline
30\% & 0.92730 \tabularnewline
40\% & 0.92361 \tabularnewline
50\% & 0.81139 \tabularnewline
60\% & 0.86707 \tabularnewline
\hline
\end{tabular}
\end{table}

In Table \ref{yale_face_table}, result of Softmax output layer is given for sample set of images, which shows that model provides correct results in most case even when objects are incomplete in an image. This proves the ability of deep learning to tolerate incompleteness of input.

\section{Conclusion}
In the presented work, we have evaluated performance of different models for image recognition. Based on the derived performance evaluations we found Inception-ResNet with highest accuracy, while keeping moderate computation requirements. Model with higher number of hidden layers improves accuracy but causes overfitting with small dataset. In order to prevent overfitting, we have used transfer learning method that efficiently reduced training time as well as overfitting without affecting accuracy of model.

\subsubsection*{Acknowledgments.}
We would like to thank Department of Computer Engineering, C. G. Patel Institute of Technology for providing us computer resources
as and when needed for training and implementing models presented in this paper.

%
%

%
%

\bibliographystyle{splncs}
\bibliography{reference}

\end{document}